\pdfoutput=1

\documentclass[11pt]{article}

\usepackage[]{acl}

\usepackage{times}
\usepackage{latexsym}

\usepackage[T1]{fontenc}

\usepackage[utf8]{inputenc}

\usepackage{microtype}

\usepackage{graphicx}
\usepackage{multirow}
\usepackage{booktabs}
\usepackage{algorithm}
\usepackage{algorithmic}
\usepackage{amssymb}
\usepackage{amsmath}
\usepackage{amsthm}

\title{Zero-shot Cross-lingual Transfer of Prompt-based Tuning \\ with a Unified Multilingual Prompt}
\newcommand\our{UniPrompt}

\author{Lianzhe Huang$\dag$\thanks{~~Contribution during internship at Microsoft.}, ~Shuming Ma$\ddag$, Dongdong Zhang$\ddag$, Furu Wei$\ddag$ and Houfeng Wang$\dag$  \\
 $\dag$MOE Key Lab of Computational Linguistics, Peking University \\
 $\ddag$Microsoft Research Asia \\
  {\tt \{hlz, wanghf\}@pku.edu.cn}\\
  {\tt \{shumma, dozhang, fuwei\}@microsoft.com} }

\begin{document}
\maketitle
\begin{abstract}
Prompt-based tuning has been proven effective for pretrained language models (PLMs). While most of the existing work focuses on the monolingual prompts, we study the multilingual prompts for multilingual PLMs, especially in the zero-shot cross-lingual setting.
To alleviate the effort of designing different prompts for multiple languages, we propose a novel model that uses a unified prompt for all languages, called \our{}.
Different from the discrete prompts and soft prompts, the unified prompt is model-based and language-agnostic.
Specifically, the unified prompt is initialized by a multilingual PLM to produce language-independent representation, after which is fused with the text input. During inference, the prompts can be pre-computed so that no extra computation cost is needed.
To collocate with the unified prompt, we propose a new initialization method for the target label word to further improve the model's transferability across languages.
Extensive experiments show that our proposed methods can significantly outperform the strong baselines across different languages. We release data and code to facilitate future research\footnote{The data and the code of this paper are available at \url{https://github.com/mojave-pku/UniPrompt}.}.

\end{abstract}
\section{Introduction}

\begin{figure}[t]
    \centering
    \includegraphics[width=0.48\textwidth]{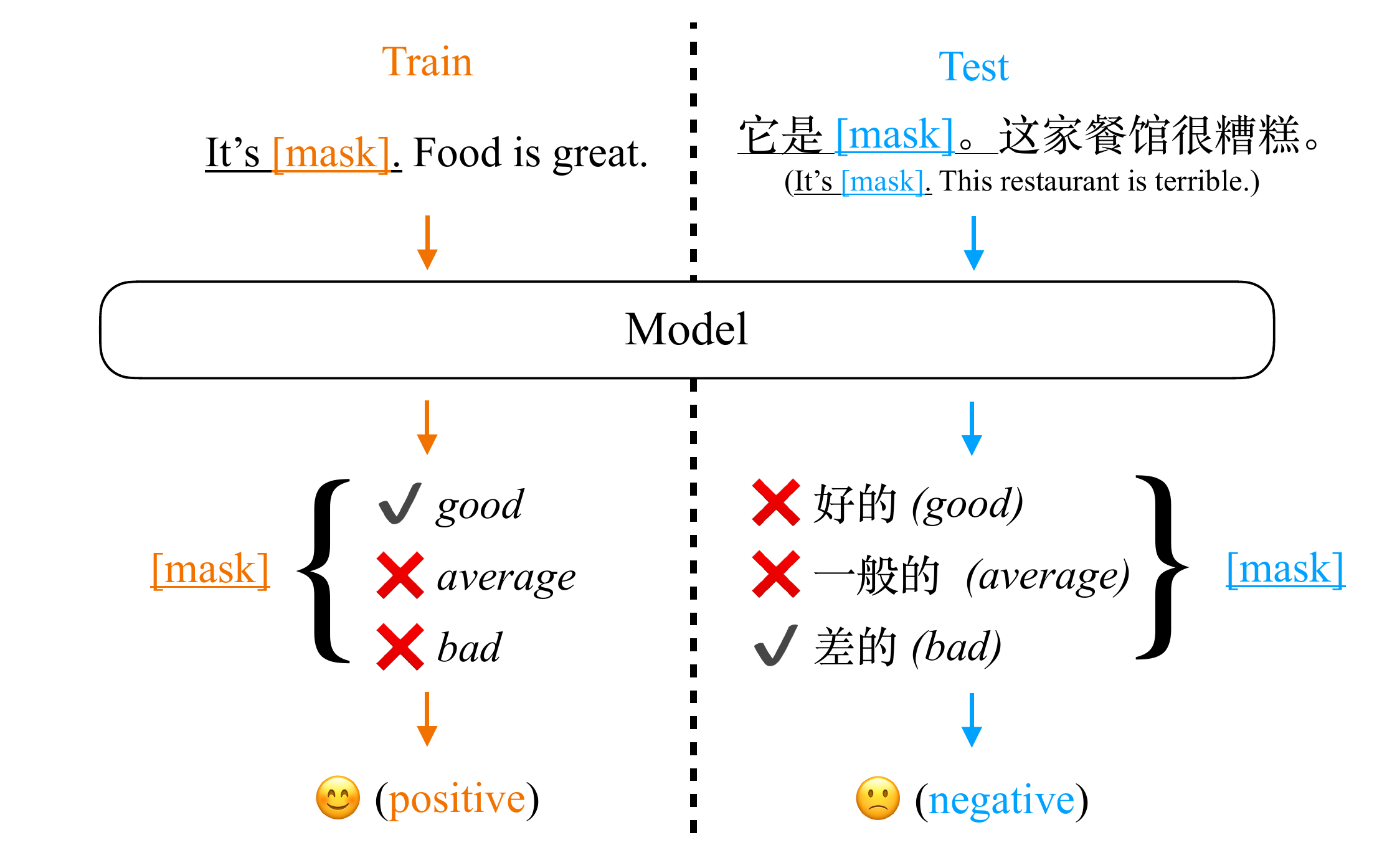} 
    \caption{An example of zero-shot cross-lingual transfer of prompt-based tuning. \underline{The underline part of the input with \texttt{[mask]} token} is template.}
    \label{fig:task_intro}
\end{figure}

Pre-trained language models (PLMs) have been proven to be successful in various downstream tasks~\cite{devlin2019bert,yang2019xlnet,conneau2020unsupervised}. Prompt-tuning is one of the effective ways to induce knowledge from PLMs to improve downstream task performance especially when the labeled data is not sufficient~\cite{brown2020language,gao2021making,le2021many,zhao2021discrete}. The essence of prompt-tuning is to precisely design task input structure so that it can imitate the pre-training procedure of PLMs and better induce the knowledge from them. For example, to classify the sentiment polarity of the source sentence ``Food is great'', a template ``It's [mask].'' is constructed before the source input where the accurate label is masked. In this way, the sentiment-related words like `good', `bad', and `average' is predicted at the masked position with probabilities on the target side, over which a verbalizer is leveraged to project the final sentiment labels.

Previously, most work on prompt-based tuning~\cite{gao2021making,zhang2021differentiable} mainly considers the monolingual prompts. However, it is not straightforward when applying to multilingual tasks due to absent multilingual prompts that heavily rely on native language experts to design both the templates and label words. An alternative way to build multilingual prompts is to machine-translate source prompts into target languages. But it is still infeasible for low-resource languages as the translation quality is hard to be guaranteed. Other work also considers using soft prompts that consist of continuous vectors.  Although it reduces the cost of building prompts for multiple languages, the mismatch between the procedures of pre-training and prompt-tuning brings many obstacles to the desired tasks, because the soft prompts never occur in the model pre-training stage.

In this work, we focus on the zero-shot cross-lingual transfer of prompt-based tuning. As shown in Figure~\ref{fig:task_intro}, the model is trained on the source language (English), while tested on the other language (Chinese). 
We explore the approaches to use a unified multilingual prompt that can transfer across languages.
We propose a novel model, called \our{}, which takes the merits of both discrete prompts and soft prompts. \our{} is model-based and language-independent. It is initialized by a multilingual PLM that takes English prompts as input and produces language-agnostic representation benefit from the transferability of multilingual PLMs. During inference, the prompts can be pre-computed so that no extra computation cost is introduced. In this way, we can alleviate the effects of prompt engineering for different languages, while reserving the ability of PLMs. To better collocate with the unified prompt, we propose a new initialization method for the label words instead of using the language model head from the PLM. This proves to further improve the model's transferability across languages.

We conducted extensive experiments on 5 target languages with different scales of data. Experimental results prove that \our{} can significantly outperform the strong baselines across different settings.
We summarize the contributions of this paper as follows:
\begin{itemize}
    \item We propose a unified prompt for zero-shot cross-lingual language understanding, which is language-independent and reserve the ability of multilingual PLMs.
    \item We propose a novel label word initialization method to improve the transferability of prompts across languages.
    \item We conduct experiments in 5 languages to prove the effectiveness of the model, and design a detailed ablation experiment to analyze the role of each module.
\end{itemize}

\section{\our{}}

\begin{table*}[t]
	\centering
	\footnotesize
\centering
\begin{tabular}{lcccccc}
\toprule
 & \textbf{En} & \textbf{De} & \textbf{Es} & \textbf{Fr} & \textbf{Ja} & \textbf{Zh} \\
\midrule 
Average characters per review & 178.8 & 207.9 & 151.3 & 159.4 & 101.4 & 51.0 \\
Number of reviews for training/development & $k\times 5$ & - & -& -& -& - \\
Number of reviews for testing & - & 5,000 & 5,000 &5,000 &5,000 &5,000  \\
\bottomrule
\end{tabular}
\caption{Statistics of MARC data used in our paper. $k$ is the number of training samples per class.}
\label{table:datasets}
\end{table*}
\subsection{Overview}

The major differences between \our{} and the existing prompt-based methods mainly lie in two parts: \textbf{template representation} and \textbf{label word initialization}.

For \textbf{template}, we use two independent encoder towers, which are the template tower and the context tower.
The template tower is to encode the prompt's template, while the context tower is for the origin text input. Both towers are initialized by the bottom layers of the multilingual PLM. After that, the representations of the template and context are concatenated as the input of the fusion tower.
The fusion tower is initialized by the top layers of multilingual PLMs.
This is motivated by the previous studies~\cite{sabet2020simalign}, which found that the lower layers of the pre-trained language model are related to language transfer, while the higher layers are related to the actual semantics.
Therefore, it can get rid of the dependency of the template on the specific language, but also retain the ability of prompts to activate the potential knowledge of PLMs.
Since the output of the prompt tower can be pre-computed before inference, the model will not introduce additional parameters or computation costs in the inference stage.

For \textbf{label words}, we use artificial tokens so that it is language-agnostic. Previous studies also have explored methods to use artificial tokens in label words~\cite{hambardzumyan2021warp}. 
Different from these works, we have a novel initialization method for the label words.
Specifically, we minimize the distance between the label words and the sentence embeddings before fine-tuning. This is achieved by taking a simple average of the sentence embeddings in the same class as the label words. In this way, the label words not only have a good starting point but also are language-independent.

\subsection{Two-tower Prompt Encoder}

As a cross-lingual unified prompt, if it directly uses the existing tokens from the vocabulary, it will be biased towards some specific languages, which will harm the cross-lingual transfer due to the gap between languages. 
To alleviate this problem, the first goal of designing a template in this task is: \textit{the template must not depend on any specific language}.
An intuitive idea to achieve this goal is to use soft prompt, which is artificial tokens that have nothing to do with specific languages. 
However, these artificial tokens: \textit{i)} will not be adequately trained due to little amount of data in few-shot scenarios; \textit{ii)} do not appear in the pre-training stage. Therefore, the goal of the prompt, which is to activate the potential knowledge of PLMs, may not be achieved. 
Given the problems of soft prompt, the second goal of designing templates can be drawn: \textit{to minimize the gaps between the pre-training and prompt-tuning.}

\begin{figure}[t]
    \centering
    \includegraphics[width=0.48\textwidth]{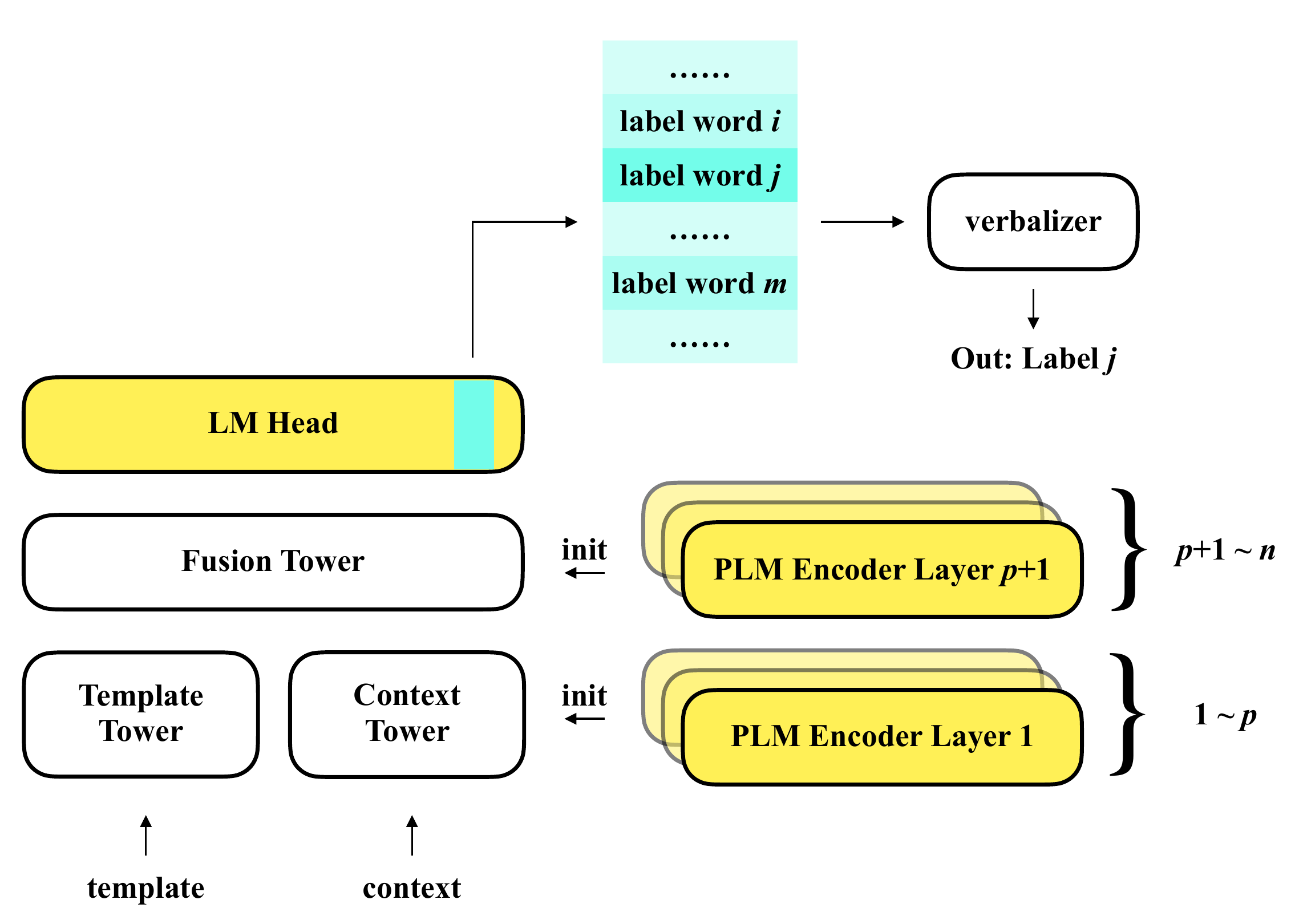} 
    \caption{Overview of Two-tower Prompt Encoder. Template Tower and Context Tower are both initialized by $1\sim p$ encoder layers of PLMs. Fusion Tower is initialized by $(p+1) \sim n$ encoder layers of PLMs. $n$ means the total number of the PLMs encoder layers.}
    \label{fig:tower}
\end{figure}

To achieve these goals, we now describe our method to model the prompts, called two-tower prompt encoder. 
The overview of the two-tower prompt encoder is shown in Figure \ref{fig:tower}.
According to the previous work, the bottom layers of PLMs encode the information related to specific language tokens/grammar, while the top layers of PLMs model the semantic information.
Therefore, we duplicate the bottom $p$ layers of PLM encoders as two independent encoder towers to encode the template and context respectively. 
Formally, we can define them as:
\begin{align}
    H^\prime_t & = {\rm TemplateTower}(X_t)\\
    H^\prime_s & = {\rm ContextTower}(X_s)
\end{align}
where $X_t, X_s$ are embeddings of template and context.

Then we concatenate the outputs of the two encoders as the input of the fusion tower which is initialized with the top $n-p$ layers of PLM:
\begin{align}
    H^\prime & = [H^\prime_t;H^\prime_s] \\
    H & = {\rm FusionTower}(H^\prime) 
\end{align}
where $n$ means the total number of the encoder layers and $[;]$ means the splicing operation. With the help of the multilingual PLM, the template tower can make the template easy to transfer across languages. 

\begin{figure}[t]
    \centering
    \includegraphics[width=0.49\textwidth]{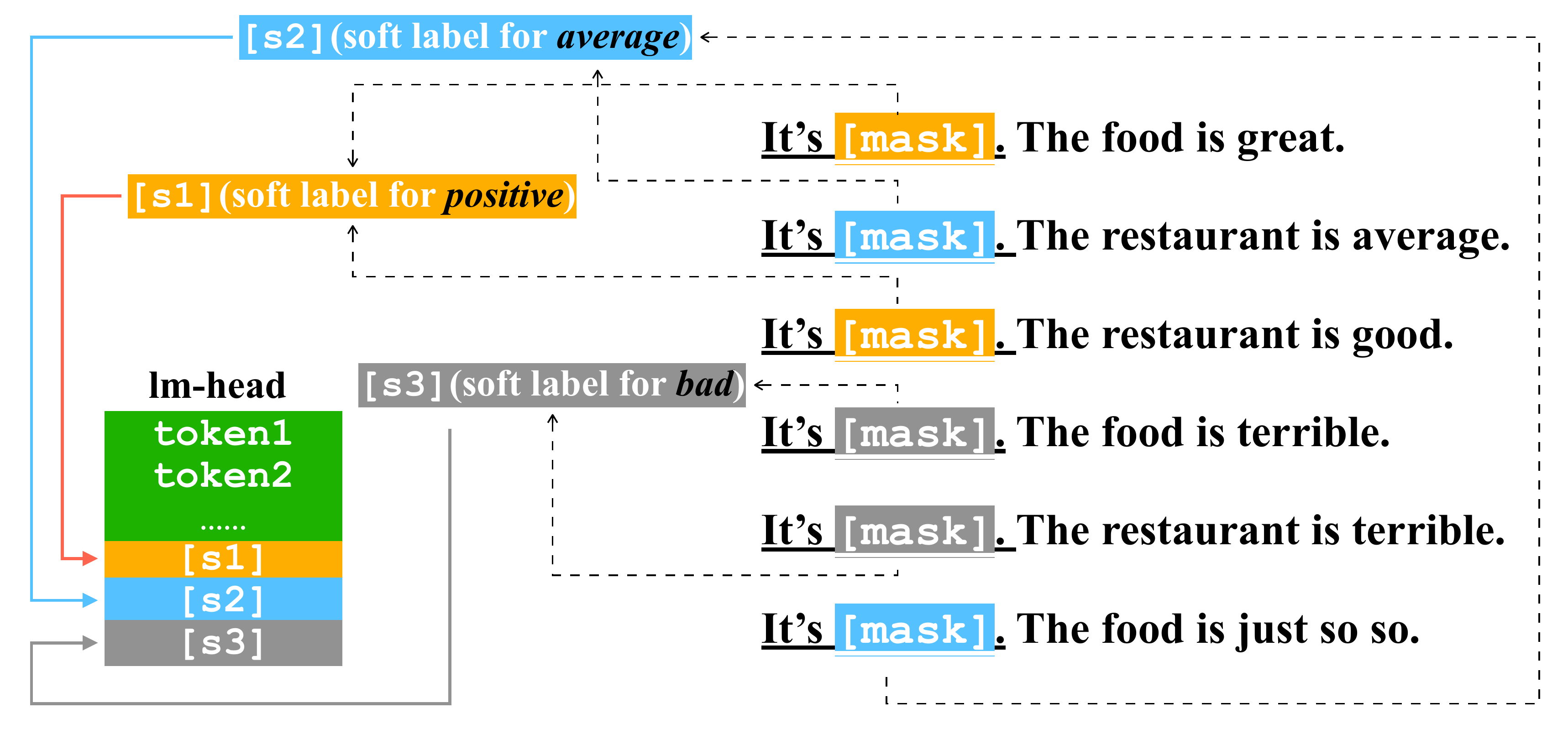} 
    \caption{An example of our proposed soft label words initialization. We first encode sentences with prompts by the original PLM, and then use the average of the representations of \texttt{[mask]} with the same label as the initialization of the soft label word.}
    \label{fig:dynamic}
\end{figure}

\subsection{Initialization of Soft Label Words}

With the two-tower prompt encoder, we are able to make the template more language-agnostic. As for label words, if we use the real tokens, they should correspond to some specific languages, which are difficult to transfer. 
Therefore, we use soft label words, i.e. artificial tokens, to achieve the goal of language independence.

To further reduce the gaps between the pre-training and fine-tuning of soft label words, we propose a novel initialization of the label words which is shown in Algorithm \ref{alg: dynamic}. And we also bring an example for this algorithm in Figure \ref{fig:dynamic}.
If we regard the output projection matrix as the word embeddings of label words, the objective of fine-tuning is to minimize the distance between the encoder outputs and the corresponding label word embedding.
Therefore, if the label word embeddings have already been close to the encoder outputs, it will be a good starting point for the models.
Motivated by this, we propose to compute the encoder outputs of all training samples, group them according to their labels, and then take a simple average of all encoder outputs in each group to initialize the label words. Note that for few-shot learning, the computation cost of pre-computing encoder outputs is small.
In this way, 
the models will have good priors to the downstream tasks while reserving the knowledge from the PLMs.

Formally, we construct soft label word $L_i$ for each label $i$, and group the training samples into $C_i$ according to their labels.
Then, we concatenate the training examples with the corresponding templates to compute the encoder outputs.
We take the average of the \texttt{[mask]} representations $h^m$ in the encoder outputs in each group to initialize the label words. The embedding $x_i$ of the label word $L_i$ can be defined as: 
\begin{equation}
    x_i = {\rm Avg}(h^m_c), c \in C_i
\end{equation}
where $\rm Avg$ means average pooling, $C_i$ is the set containing the training cases with label $i$. 
\begin{algorithm}[t]
  \caption{Initialization of Soft Label Words }
  \label{alg: dynamic}
\begin{algorithmic}[1]
  \STATE {\bfseries Input:} original pre-trained language model $\boldsymbol{\theta}^{0}$, all training cases $C_i$ with label $i$, prompt with \texttt{[mask]} token $p$
  \FOR{each case $c_j$ in $C_i$}
    \STATE form the prompt input $c^\prime_j$ for encoding: $c^\prime_j \leftarrow p + c_j$
    \STATE encode the sequence $c^\prime_j$ without gradients: $H_j \leftarrow \boldsymbol{\theta}^{0}(c^\prime_j)$
    \STATE get the representation of \texttt{[mask]} token $h^m_j$ from $H_j$
  \ENDFOR
  \STATE average all the $h^m$ as the representation $x_i$ of the soft label word for label $i$: $x_i \leftarrow  {\rm Avg}({h^m_j}), {j \in C_i}$
  \RETURN $x_i$
\end{algorithmic}
\end{algorithm}

\subsection{Training}
Similar to the previous prompt-based tuning method, we use the distribution probability of label words for classification:
\begin{equation}
    {\rm logit_y} = \frac{\exp(\mathbf{W_{lh}^y}h^m)}{\sum_{y^\prime \in \mathcal{Y}}\exp(\mathbf{W_{lh}^{y\prime}}h^{m})} \label{eq:logits}
\end{equation}
where $\mathcal{Y}$ is the set of all labels, $\mathbf{W_{lh}^i}$ is the parameters corresponding to the label $i$ from the output projection matrix (i.e. label word embeddings).

The loss function $\mathcal{L}$ in our model is the cross-entropy loss, which can be defined as: 
\begin{align}
    \mathcal{L} &= -g_y\log ({\rm logit_y}),
\end{align}
where $g_y$ is the ``one-hot vector'' of gold labels.
\\

The overall workflow of our proposed \our{} is shown in Algorithm \ref{alg: overall}.
\begin{algorithm}[t]
  \caption{Overall Workflow of \our{} }
  \label{alg: overall}
\begin{algorithmic}[1]
  \STATE {\bfseries Input:} pre-trained language model $\boldsymbol{\theta}^{0}$, prompt $p$, cases $c$ with label
  \FOR{each label $i$}
    \STATE group all the cases with label $i$ as $C_i$
    \STATE initialize soft label word $x_i \leftarrow $  Algorithm \ref{alg: dynamic} ($\boldsymbol{\theta}^{0}, p, C_i$)
  \ENDFOR
  \FOR{each training case $c_j$}
    \STATE prompt and $c_j$ are sent into the two-tower prompt encoder for encoding, respectively: $H^j_t \leftarrow {\rm TemplateTower}(p)$,  $H^j_c \leftarrow {\rm ContextTower}(c_j)$
    \STATE The two vectors $H_t^{j\prime}, H_t^{j\prime}$ are stitched together and fed into the fusion tower to get the final representation $H^j \leftarrow {\rm FusionTower}(H_t^{j\prime};H_t^{j\prime})$
    \STATE get the label by the prediction result of maskLM task : $y \leftarrow {\rm maskLM}(h_j^m)$
  \ENDFOR
  
\end{algorithmic}
\end{algorithm}

\section{Experiments}
\begin{table*}[t]
\centering
\scalebox{0.85}{
\begin{tabular}{clcccccc}
\toprule
\textbf{$k$} & \textbf{Model} & \textbf{De} & \textbf{Es} & \textbf{Fr} & \textbf{Ja} & \textbf{Zh} & \textbf{Average} \\
\midrule

\multirow{5}*{4} 
& Vanilla Finetune &  $27.94 \pm 8.38$ & $ 26.80 \pm 5.96$ & $27.38 \pm 6.48$ & $25.86 \pm 5.08$ & $25.95 \pm 6.51$ & $ 26.79$ \\
& Translation Prompt & $30.76 \pm 4.48$ & $31.70 \pm 3.46$ & $26.77 \pm 3.49$ & $27.46 \pm 4.52$ & $21.56 \pm 2.30$ & $27.65$ \\
& English Prompt &  $33.52 \pm 5.14$ & $33.01 \pm 3.61$ & $33.32 \pm 3.54$ & $32.23 \pm 2.25$ & $30.97 \pm 4.87$ & $\mathbf{32.61}$ \\
& Soft Prompt & $29.26 \pm 9.28$ & $30.74 \pm 3.22$ & $31.10 \pm 5.72$ & $28.96 \pm 0.76$ & $28.26 \pm 3.56$ & $29.66$ \\
& \bf \our{} & $31.70 \pm 5.42$ & $30.79 \pm 5.73$ & $30.97 \pm 6.29$ & $30.21 \pm 6.37$ & $28.70 \pm 4.40$ & $30.47$ \\
\midrule

\multirow{5}*{8} 
& Vanilla Finetune &  $31.95 \pm 7.07$ & $29.74 \pm 7.38$ & $31.79 \pm 6.41$ & $28.70 \pm 5.30$ & $29.67 \pm 7.25$ & $30.37$ \\
& Translation Prompt & $33.04 \pm 1.04$ & $35.22 \pm 1.16$ & $28.93 \pm 1.51$ & $30.10 \pm 2.38$ & $23.62 \pm 2.80$ & $30.18$ \\
& English Prompt &  $36.96 \pm 1.94$ & $36.12 \pm 0.98$ & $36.70 \pm 1.38$ & $33.95 \pm 4.21$ & $32.97 \pm 2.71$ & $35.34$ \\
& Soft Prompt & $33.30 \pm 3.22$ & $32.88 \pm 2.34$ & $33.28 \pm 2.44$ & $30.05 \pm 3.87$ & $29.46 \pm 4.22$ & $31.79$ \\
& \bf \our{} & $38.58 \pm 2.96$ & $37.68 \pm 3.38$ & $37.88 \pm 3.80$ & $35.72 \pm 5.60$ & $34.57 \pm 5.69$ & $\mathbf{36.89}$ \\
\midrule

\multirow{5}*{16} 
& Vanilla Finetune &  $40.10 \pm 5.46 $ & $38.37 \pm 3.55 $ & $38.73 \pm 3.59 $ & $36.20 \pm 4.82 $ & $35.94 \pm 5.74 $ & $37.87$\\
& Translation Prompt & $36.62 \pm 1.72$ & $36.88 \pm 1.14$ & $32.48 \pm 2.90$ & $31.98 \pm 1.42$ & $24.20 \pm 2.14$ & $32.43$ \\
& English Prompt &  $39.13 \pm 3.63$ & $37.84 \pm 3.34$ & $38.58 \pm 1.90$ & $35.90 \pm 4.44$ & $35.05 \pm 5.59$ & $37.30$ \\
& Soft Prompt & $37.25 \pm 3.57$ & $34.96 \pm 2.74$ & $35.18 \pm 3.20$ & $34.64 \pm 2.76$ & $34.20 \pm 4.44$ & $35.24$ \\
& \bf \our{} & $43.53 \pm 5.11$ & $41.43 \pm 4.39$ & $41.71 \pm 5.21$ & $39.55 \pm 4.41$ & $38.62 \pm 2.82$ & $\mathbf{40.97}$ \\
\midrule

\multirow{5}*{32} 
& Vanilla Finetune &  $45.48 \pm 2.74$ & $42.09 \pm 4.21$ & $43.14 \pm 2.42$ & $40.86 \pm 4.74$ & $41.39 \pm 1.61$ & $42.59$ \\
& Translation Prompt & $39.29 \pm 3.25$ & $38.46 \pm 1.96$ & $34.75 \pm 4.23$ & $34.76 \pm 2.52$ & $26.88 \pm 5.64$ & $34.83$  \\
& English Prompt & $42.04 \pm 2.32$ & $40.39 \pm 1.51$ & $41.33 \pm 2.39$ & $39.06 \pm 2.34$ & $37.72 \pm 3.60$ & $40.11$ \\
& Soft Prompt & $40.58 \pm 1.48$ & $38.29 \pm 2.05$ & $39.50 \pm 2.28$ & $38.80 \pm 1.90$ & $35.90 \pm 4.94$ & $38.61$ \\
& \bf \our{} & $49.29 \pm 2.11$ & $46.74 \pm 1.28$ & $47.47 \pm 1.03$ & $45.94 \pm 1.98$ & $43.62 \pm 1.16$ & $\mathbf{46.61}$ \\
\midrule

\multirow{5}*{64} 
& Vanilla Finetune &  $49.85 \pm 2.73$ & $45.74 \pm 3.18$ & $47.72 \pm 2.76$ & $44.68 \pm 3.80$ & $43.84 \pm 1.76$ & $46.37$ \\
& Translation Prompt & $41.70 \pm 3.72$ & $40.93 \pm 1.85$ & $37.24 \pm 3.54$ & $36.75 \pm 1.03$ & $29.72 \pm 2.06$ & $37.27$ \\
& English Prompt & $45.63 \pm 4.25$ & $43.41 \pm 4.33$ & $44.18 \pm 3.30$ & $41.43 \pm 3.65$ & $39.63 \pm 2.13$ & $42.86$ \\
& Soft Prompt & $44.49 \pm 2.97$ & $40.73 \pm 3.35$ & $42.15 \pm 3.79$ & $41.91 \pm 3.09$ & $40.36 \pm 4.36$ & $41.93$ \\
& \bf \our{} & $51.75 \pm 1.57$ & $47.64 \pm 1.70$ & $49.54 \pm 0.98$ & $45.99 \pm 1.75$ & $45.55 \pm 3.13$ & $\mathbf{48.09}$ \\
\midrule

\multirow{5}*{128} 
& Vanilla Finetune &  $51.84 \pm 1.82$ & $49.09 \pm 1.17$ & $49.52 \pm 1.20$ & $47.74 \pm 3.12$ & $45.81 \pm 1.73$ & $48.80$ \\
& Translation Prompt & $42.99 \pm 2.99$ & $41.16 \pm 1.52$ & $36.78 \pm 1.58$ & $37.12 \pm 1.92$ & $29.46 \pm 2.96$ & $37.50$ \\
& English Prompt &  $49.36 \pm 2.96$ & $45.70 \pm 2.18$ & $45.98 \pm 3.42$ & $44.12 \pm 2.84$ & $43.97 \pm 1.31$ & $45.83$ \\
& Soft Prompt & $48.06 \pm 1.60$ & $42.64 \pm 3.94$ & $43.41 \pm 4.01$ & $45.22 \pm 1.64$ & $44.19 \pm 1.43$ & $44.70$ \\
& \bf \our{} & $53.18 \pm 1.28$ & $49.74 \pm 0.98$ & $50.22 \pm 0.68$ & $48.48 \pm 1.30$ & $46.47 \pm 1.31$ & $\mathbf{49.62}$ \\
\midrule

\multirow{5}*{256} 
& Vanilla Finetune &  $53.06 \pm 1.24$ & $50.05 \pm 1.57$ & $50.47 \pm 1.33$ & $47.93 \pm 3.63$ & $46.40 \pm 2.72$ & $49.58$ \\
& Translation Prompt & $46.44 \pm 1.04$ & $43.06 \pm 1.30$ & $38.76 \pm 1.44$ & $38.49 \pm 1.79$ & $29.40 \pm 3.62$ & $39.23$ \\
& English Prompt & $51.66 \pm 0.60$ & $47.90 \pm 2.52$ & $48.18 \pm 2.32$ & $47.32 \pm 1.70$ & $44.77 \pm 1.27$ & $47.97$ \\
& Soft Prompt & $50.81 \pm 0.83$ & $44.92 \pm 1.76$ & $45.29 \pm 1.97$ & $47.36 \pm 1.56$ & $44.24 \pm 2.64$ & $46.52$ \\
& \bf \our{} & $54.36 \pm 1.16$ & $51.13 \pm 0.83$ & $51.56 \pm 0.86$ & $49.66 \pm 1.64$ & $47.57 \pm 1.29$ & $\mathbf{50.86}$ \\

\bottomrule
\end{tabular}
}
\caption{Main results. $k$ is the number of training samples per class (i.e. $k$-shot).}

\label{table:main_results}
\end{table*}

\subsection{Datasets}
We choose the Multilingual Amazon Reviews Corpus (MARC)\footnote{https://registry.opendata.aws/amazon-reviews-ml/}~\cite{keung2020multilingual} for experiments, which is a large-scale multilingual text classification dataset with the licence provided by Amazon\footnote{https://github.com/awslabs/open-data-docs/blob/main/docs/amazon-reviews-ml/license.txt}. The MARC dataset is available in 6 languages including English, German, French, Spanish, Japanese, and Chinese. 
The goal of this dataset is to predict the star rating given by the reviewer to the product based on their product reviews (from 1 to 5 stars, the higher the star rating, the more satisfied they are).

In the MARC dataset, the number of samples in each category is exactly the same, and we follow their settings to take the same samples for each category to form few-shot training and development sets.
In our experiments, we randomly sample $k$ cases from each category, $k\times 5$ cases in total, to form new training and development sets, and the test set remains unchanged.
An overview of the dataset is shown in Table \ref{table:datasets}, and some statistics are directly taken from \citet{keung2020multilingual}. Our source language is English, which is the language used for the training and development sets.
The target languages, which include the remaining 5 languages, are used for the test set.

The task and dataset we used are representative and challenging.
Text classification is one of the fundamental problems for NLP. It also proves to be a good test bed for few-shot learning according to the previous work.
The MARC dataset used in this work is challenging, especially for the multilingual few-shot scenarios. According to our experiments, vanilla fine-tuning only gets an average accuracy of $26.79$ in the 4-shot setting. Therefore, we believe the benchmark is sound.

\subsection{Experimental Setup}
Our method is based on \texttt{XLM-RoBERTa-base} model~\cite{conneau2020unsupervised}, which is a widely used multilingual pretrained language model.
We implement our model with HuggingFace \texttt{Transformers} \cite{wolf-etal-2020-transformers} and code released by \citet{gao2021making}. 
We optimize our models with a learning rate of {1e-5}. The batch size is set to 8. We train each model for 1000 steps and evaluate per 100 steps, the best checkpoint is used for the final prediction.
The number of layers used for prompt and context towers is set to 9. 
The max sequence length of the model is set to 512. 
For each experiment reported in the paper, we use 5 different random seeds to sample 5 different  few-shot training/development dataset from the original one. We run the model with the same random seeds as the one for dataset sampling and report the average results.

We also want to introduce the number of trainable parameters. Since we do not freeze the parameters during training, the trainable parameter number of the baseline is the total parameter number of \texttt{XLM-RoBERTa-base}. During the training of our model, there is an additional number of parameters from the template tower. The specific number is related to the number of layers ($L$) of the template tower, which will bring an additional parameter number of $L*p$, where $p$ is the number of parameters for each layer. During inference, our model uses a template tower output cache, and the number of parameters is the same as the baseline.

\subsection{Baselines}
We compare the model with the following baseline models, all parameters in the baseline models are not frozen:\\
\textbf{Vanilla Finetune} add a task-dependent linear layer after the pretrained language model for classification~\cite{devlin2019bert}.\\
\textbf{Translation Prompt} proposed by~\citet{zhao2021discrete}, which uses the source language prompt for training and translates the prompt into the target language by machine translation model for testing. \\
\textbf{English Prompt} proposed by~\citet{lin2021few}, which trains and tests by prompts in the source language (English).\\
\textbf{Soft Prompt} uses artificial tokens instead of discrete tokens as templates, the label words are still in the source language.

The baseline models above are all implemented by us
on the same codebase, initialized by the same pre-trained language model, and the hyper-parameters, including \texttt{max\_steps}, \texttt{eval\_steps}, \texttt{batch\_size}, \texttt{learning\_rate}, \texttt{max\_seq\_length}, and so on, are consistent. All experiments are performed on the same computing cluster with the same docker image.

\subsection{Main Results}
The main experimental results are shown in Table \ref{table:main_results}. As can be seen from the experimental results, our model outperforms all the listed baseline models at all data scales except slightly lower than \textbf{English Prompt} in the case of a very small amount of data ($k=4$).

From the perspective of data scales, our model performs very well on medium data sizes ($k=16,32,64$), with an average 2\% higher accuracy than the strongest baseline. Especially when $k=32$, the accuracy is more than 4\% higher than the strongest baseline, which fully proves the ability of our model in few-shot cross-lingual transfer. As the size of the data continues to increase, the model leads by a smaller margin. But even if the data scale reaches $k=256$, the accuracy of our model is still at least 1\% higher than all other baselines.

Next, we discuss the comparison with each baseline separately. First, our model outperforms \textbf{Vanilla Finetune} models on all languages and data scales. We believe the reasons for the worse performance of \textbf{Vanilla Finetune} include: \textit{i)} in the vanilla fine-tune model, a task-related linear layer is added on the top of PLMs. This layer is randomly initialized and requires more training data to be fully trained, which results in failure on low-resource tasks. \textit{ii)} failing to exploit the latent knowledge in large-scale unlabeled corpus like prompt.

Second, our model also performs better than the \textbf{Translation Prompt} model~\cite{zhao2021discrete}. 
Converting the prompt directly using a machine translation model is indeed an intuitive and less expensive method. But there are also some problems. 
\textit{i)} the model will be limited by the machine translation model, potentially causing error propagation. \textit{ii)} since the translated prompt model has never been seen during training, the model cannot be properly fine-tuned according to the dataset situation, which may also lead to performance loss. 

Next, we discuss \textbf{English Prompt}, which directly use the prompt from the source language. The \textbf{English Prompts} fits the training data which is also in English, so when the data scale is very small ($k=4$), this method achieves the best results. But as the amount of data gets slightly larger ($k=8$, which is still a very small scale), the performance of \textbf{English Prompt} is not as good as \textbf{\our{}}. The key point of the task in this paper is to enhance the cross-language transfer ability of the model. Since the PLMs are not trained by cross-language splicing texts in the pre-training phase, so when an English prompt is combined with the context in another language in the testing phase of cross-lingual transfer, there will be gaps with the pre-training phase, which results in performance loss.

Finally, our model also has better performance than \textbf{Soft Prompt}. 
Although the soft prompt has nothing to do with the specific language and is consistent during training and testing. But 
\textit{i)} it has not appeared in the pre-training stage, so it may be difficult to activate the latent knowledge in the pre-training stage. And 
\textit{ii)} in low-resource scenarios, the completely randomly initialized soft prompt cannot be fully trained.

We note that the standard deviation of the experimental results is relatively large due to the small scale data of few-shot settings, which may lead to confusion about whether the performance gain is significant, especially the performance difference between the vanilla fine-tune and our model on some data scales. For this we selected the original experimental results of vanilla fine-tune and our UniPrompt in all 5 languages (de/es/fr/ja/zh) when $k=16$ for the statistical test, the results indicate that the performance difference between our method and vanilla fine-tune is statistically significant.

\section{Analysis}
\begin{figure}[t]
    \centering
    \includegraphics[width=0.48\textwidth]{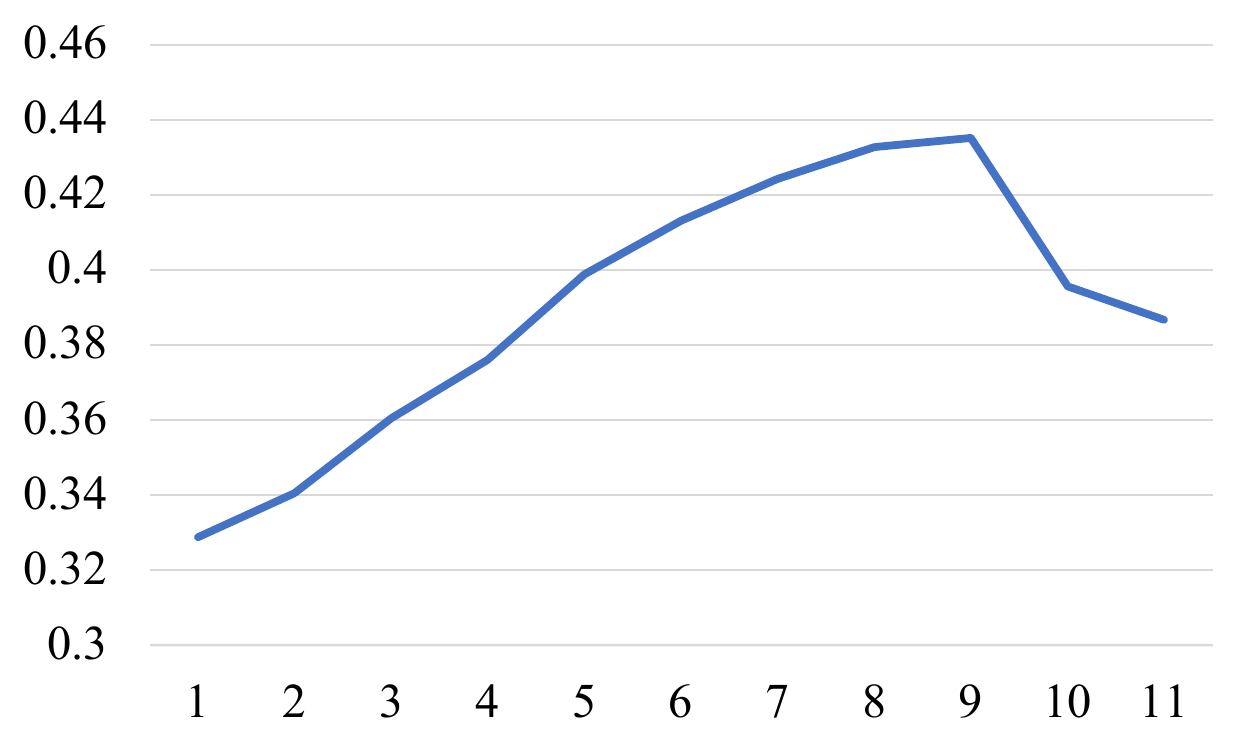} 
    \caption{The effect of different number of template tower and context tower's layers. Except for the number of layers, the other settings are the same as the main experiment.}
    \label{fig:layers}
\end{figure}
\begin{table*}[t]
\centering
\scalebox{0.87}{
\begin{tabular}{lcccccc}
\toprule
\textbf{Initialization} & \textbf{De} & \textbf{Es} & \textbf{Fr} & \textbf{Ja} & \textbf{Zh} & \textbf{Average} \\
\midrule
Random Initialization &  $40.24 \pm 6.48$ & $39.60 \pm 3.32$ & $40.25 \pm 5.45$ & $37.80 \pm 4.34$ & $37.51 \pm 3.71$ & $39.08$ \\
RoBERTa-base &  $38.96 \pm 4.56$ & $37.85 \pm 3.75$ & $37.90 \pm 2.94$ & $36.14 \pm 3.10$ & $36.05 \pm 5.13$ & $37.38$ \\
XLM-RoBERTa-base & $43.53 \pm 5.11$ & $41.43 \pm 4.39$ & $41.71 \pm 5.21$ & $39.55 \pm 4.41$ & $38.62 \pm 2.82$ & $40.97$ \\
\bottomrule
\end{tabular}
}
\caption{Results of analysis on the template tower initialization.}

\label{table:state_dict}
\end{table*}
\begin{table}[t]
    \centering
    \includegraphics[width=0.49\textwidth]{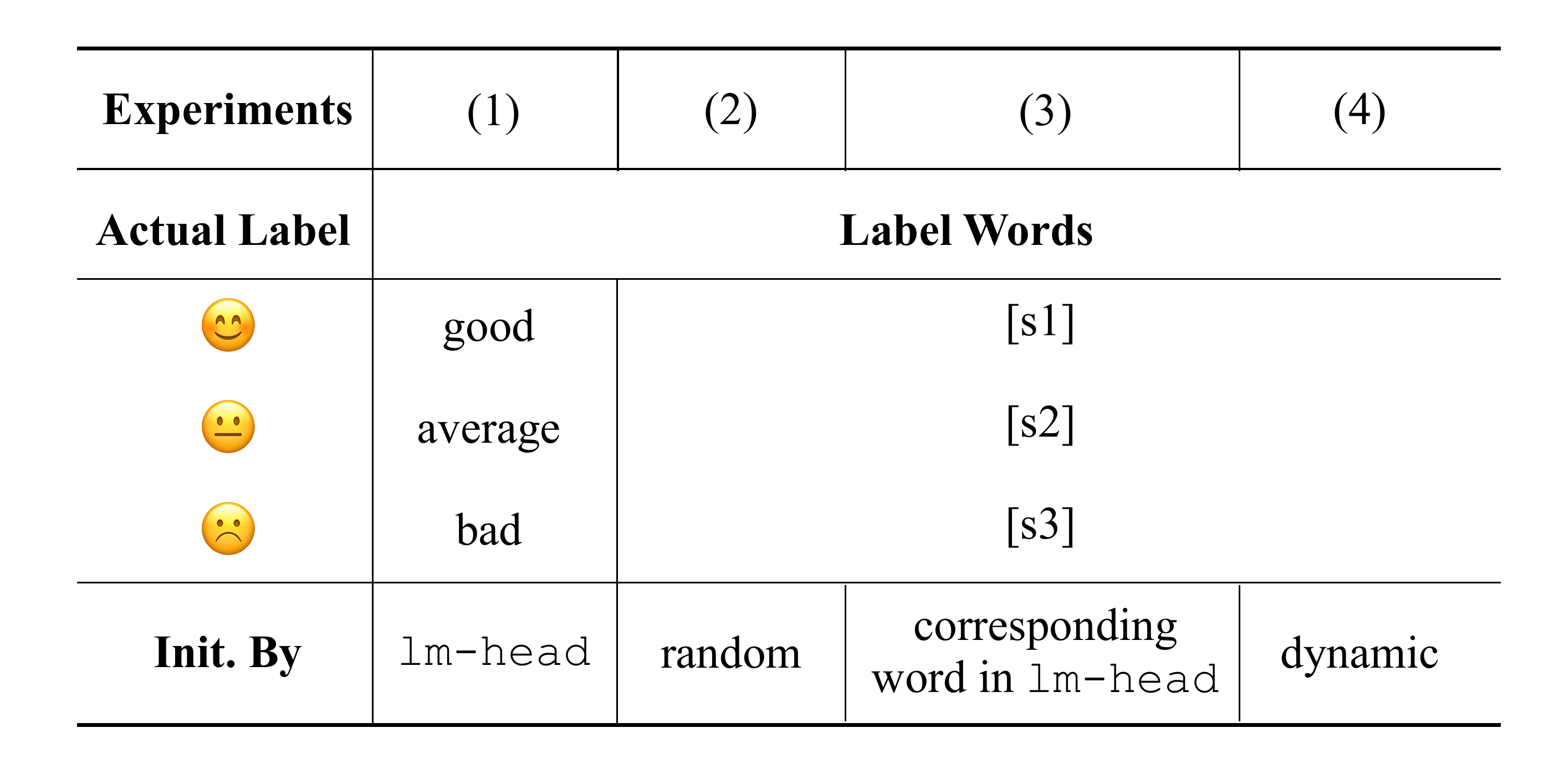} 
    \caption{The comparison of the settings of ablation study.}
    \label{table:ablation}
\end{table}
\begin{table*}[t]
\centering
\scalebox{0.9}{
\begin{tabular}{lcccccc}
\toprule
\textbf{Label Words} & \textbf{De} & \textbf{Es} & \textbf{Fr} & \textbf{Ja} & \textbf{Zh} & \textbf{Average} \\
\midrule
(1) Discrete Labels &  $38.94 \pm 2.46$ & $37.84 \pm 1.26$ & $37.57 \pm 1.87$ & $36.80 \pm 3.64$ & $34.78 \pm 4.02$ & $37.19$ \\
(2) Soft Labels & $40.64 \pm 2.20$ & $38.90 \pm 0.98$ & $39.97 \pm 2.27$ & $37.51 \pm 2.79$ & $37.24 \pm 2.10$ & $38.85$ \\
(3) (2) + PLM Init. &   $42.22 \pm 4.28$ & $40.51 \pm 2.03$ & $40.60 \pm 3.02$ & $37.94 \pm 3.12$ & $37.81 \pm 2.51$ & $39.82$ \\
(4) (2) + our Init. &  $43.53 \pm 5.11$ & $41.43 \pm 4.39$ & $41.71 \pm 5.21$ & $39.55 \pm 4.41$ & $38.62 \pm 2.82$ & $40.97$ \\
\bottomrule
\end{tabular}
}
\caption{Ablation study on the label words.}

\label{table:label_words}
\end{table*}
In this section, we will analyze the model in detail to verify the effectiveness of each module of the model. 
All experimental setups in this section are identical to the main experiments unless otherwise stated, and all experiments are based on 16-shot data.
\subsection{Discussion on Two-tower Prompt Encoder}
We first discuss the Two-tower Prompt Encoder. According to its setting, we discuss the effects of the number of layers and the pretrained models on model performance separately.
\subsubsection{Number of Layers}
As discussed above, the reason why the Two-tower Prompt Encoder works is that it splits the bottom encoders of PLMs, which are considered syntax-related, into two separate encoder towers, making the prompt free from language-specific dependencies.
At the same time, when entering the top encoder of PLMs, which is considered to be semantically related, the two representations are fused, thereby stimulating the potential knowledge of PLMs in the pre-training stage.

A key question about the Two-tower Prompt Encoder is the \textbf{dividing line} between the top and bottom layers of PLMs. In response to this, we conducted experiments on the \texttt{en->de} data, and the experimental results are shown in the Figure \ref{fig:layers}.
From the experimental results, we can see that the \textbf{dividing line} we expect is at 9 out of 12 layers, about 75\% of the encoder layers of the PLMs.
Before the dividing line, as the number of the independent lower and syntax-related encoder layers increases, the effect of decoupling the prompt from the specific language becomes better. Therefore, the performance of the model is gradually improved.
After the dividing line, with the increase of the number of independent layers, although the ability to decouple with specific languages becomes stronger, the number of layers left for the fusion of template and context becomes less, and too little fusion limits the capacity prompt for activating the latent knowledge during the pre-training phase, so model performance gradually decreases.

\subsubsection{Pretrained Models}

Another point worth discussing about the Two-tower Prompt Encoder is the pretrained language models for template tower initialization.
In our experiments, the template tower is initialized from the corresponding layers of the encoders of multilingual PLMs like the context/fusion tower.
Therefore, we analyze whether the improvement is brought by the transferability from the multilingual PLM.
Since in the original model, the prompt is initialized with the source language, that is, the English prompt. An intuitive idea is to use the English monolingual PLMs for template tower initialization. Compared with multilingual PLMs, English PLMs do not have the ability of cross-lingual transfer, which helps us ablate the effects of transferability. In addition, we also compare with random initialization, which should not benefit from the PLMs. The results are shown in Table~\ref{table:state_dict}.

From the experimental results, it can be seen that using multilingual PLMs, which have cross-lingual transferability, achieve the best results. And there is a notable phenomenon that even if the template is based on the source language, initializing the template tower by the monolingual PLM RoBERTa~\cite{liu2019roberta} performs worse than random initialization. This indicates that the cross-lingual transferability is much more important than the effect of the PLMs itself for our method.

\subsection{Discussion on Label Words}
We also conduct the ablation study to verify the effect of our soft label words initialization method. 
The comparison of the settings of the 4 groups of experiments is shown in Table \ref{table:ablation}.
The experimental results are shown in Table~\ref{table:label_words}.
First, we compare soft label words with discrete label words. The experimental results show that removing soft labels leads to a significant drop in performance.
This illustrates the necessity of decoupling label words from specific languages in cross-lingual tasks. There will be gaps in the pre-training stage by using discrete tokens that depend on specific languages as label words during cross-language transfer, and the ability of prompts to activate knowledge in the pre-training stage will be weakened, resulting in performance loss.

Then, we compare the models with and without initialization. It shows that the initialization of label words results in a significant gain, proving our motivation that initialization is important for label words. We also compare our initialization method with the original PLM initialization. Compared (3) with (4), it proves that our initialization method is effective to improve cross-lingual performance. This is because our initialization method for the label words can reduce the gaps between the pre-training and the fine-tuning.

\section{Related Work}

\subsection{Prompt-based tuning}
The proposal of GPT-3 inspired the research on prompt~\cite{brown2020language}.
The key to prompt tuning is to reasonably imitate the pre-training process of the PLM model, so as to maximize the implicit knowledge learned by the model from the large-scale unlabeled corpus.
Most of the existing template-based research focuses on how to design or search for the best template suitable for downstream tasks~\cite{le2021many,zhang2021differentiable,li2021prefix}, but does not focus on optimization from aspects such as model parameters or structure.
As for label words, almost all models are still using discrete tokens as label words. 
\citet{hambardzumyan2021warp} proposed to use artificial tokens as label words, but they used randomly initialized label words and did not consider finding a better initialization for them, which may bring performance loss.

\subsection{Zero-shot Cross-lingual Transfer}
At a time when labeling resources are expensive, the research on Zero-shot Cross-lingual Text Classification is quite valuable. 
Past research in this area is usually based on cross-task transfer learning, that is, the model is first trained on a dataset of resource-rich tasks, and then fine-tuned on specific low-resource downstream tasks \cite{pruksachatkun2020intermediate,zhao2021closer}.
As research on prompts progresses, prompts have been found to perform well on low-resource tasks~\cite{brown2020language,liu2021pre}. 
But most of the research on prompt-based text classification is monolingual~\cite{gao2021making,liu2021p}.
And there are some problems with the few multilingual studies.
\citet{zhao2021discrete} first uses a prompt-based approach on this task. They propose a hard prompt based on machine translation, but this approach relies on machine translation models and may introduce additional errors. They also proposed to use soft prompts. Although it can be decoupled from the specific language, there are still gaps between the randomly initialized soft prompt and the pre-training stage.
\citet{lin2021few} proposes to use English prompts with non-English examples, but they did not consider decoupling the specific language from the view of the model structure, and still used discrete tokens as label words.
\citet{winata2021language} performs few-shot multilingual multi-class classification without updating parameters by applying binary prediction and considering the confidence scores of boolean tokens. Although freezing parameters has the advantage of the model training cost, the model cannot be fine-tuned according to the actual task, which may lead to performance loss. 
\section{Conclusion}
In this paper, we propose a new prompt-based tuning method for zero-shot cross-lingual text classification. 
For the two key elements of the prompt, we respectively give solutions under this task setting.
For templates, we use a two-tower prompt encoder for encoding, which not only decouples specific languages but also preserves the ability of prompts to activate the latent knowledge of the language model.
For label words, we use soft label words and dynamic initialization methods, which also achieve the goal of decoupling specific languages.
The experimental results prove the utility of our model, and we also design experiments to carry out detailed analysis of the settings of our model. 
\section*{Limitations}

\paragraph{Our UniPrompt is more suitable for low resource scenarios.} With the growth of the data scale, the advantages of UniPrompt become minor. This is also verified by the existing prompt-based methods. Prompt can stimulate the potential knowledge of PLM in the pre-training stage, which may be general and may not match the domain knowledge of downstream tasks. In the case of a small data scale, this general knowledge can greatly help the model to judge with limited domain knowledge. With the growth of the data scale, the model can summarize the corresponding knowledge which is adapted to the task from the domain data, and the importance of general knowledge brought by prompt decreases relatively.
\paragraph{Currently, our method is only applicable to natural language understanding tasks.} This is determined by the selected PLM model, type of prompt, and model structure. We believe that some of the ideas in this paper can be used in natural language generation, which remains to be further investigated by subsequent research.

\section*{Acknowledgments}
The work is supported by National Natural Science Foundation of China under Grant No.62036001 and PKU-Baidu Fund (No. 2020BD021). The corresponding author of this paper is Houfeng Wang.

\bibliography{anthology,custom}

\end{document}